\documentclass{article}

\usepackage[nonatbib]{neurips_2020}

\usepackage[utf8]{inputenc} % allow utf-8 input
\usepackage[T1]{fontenc}    % use 8-bit T1 fonts
\usepackage{hyperref}       % hyperlinks
\usepackage{url}            % simple URL typesetting
\usepackage{booktabs}       % professional-quality tables
\usepackage{amsfonts}       % blackboard math symbols
\usepackage{nicefrac}       % compact symbols for 1/2, etc.
\usepackage{microtype}      % microtypography

\title{Compressed Object Detection}

\author{%
  Gedeon M.~Hippocampus\thanks{https://gedeonmuhawenayo.github.io/} \\
  Department of Machine Intelligence\\
  African Institute for Mathematical Sciences\\
  Kigali, Rwanda \\
  \texttt{gmuhawenayo@aimsammi.org} \\
}

\begin{document}

\maketitle

\begin{abstract}
  Deep learning approaches have achieved unprecedented performance in visual recognition tasks such as object detection and pose estimation. However, state-of-the-art models have millions of parameters represented as floats which make them computational expensive and constrain their deployment on hardware such as mobile phones and IoT nodes. Most commonly, activations of deep neural networks tend to be sparse thus proving that models are over parametrized with redundant neurons. Model compression techniques, such as pruning and quantization, have recently shown promising results by improving model complexity with little loss in performance. In this work, we extended pruning, a compression technique which discards unnecessary model connections, and weight sharing techniques for the task of object detection. With our approach we are able to compress a state-of-the-art object detection model by 30.0\% without a loss in performance. We also show that our compressed model can be easily initialized with existing pre-trained weights, and thus is able to fully utilize published state-of-the-art model zoos. 
\end{abstract}

\section{Introduction}
Deep neural networks are computationally expensive. Their memory and compute complexity limit their deployment on edge devices and make them unsuitable for applications with strict latency requirements \cite{BNN_survey}. The progress in VR, AR, IoT and smart wearable devices create opportunities for researchers to tackle the challenges of deploying deep neural networks on devices with constrained memory, CPU, bandwidth and energy \cite{objects_detection_survey,liu2018rethinking}.
To this end, compression techniques have shown promising results in reducing the memory and time requirements of deep neural networks. Pruning reduces the model's complexity by removing unnecessary elements in their structures at different levels  \cite{pas2020modeling}. 
Quantization converts a model to use reduced precision integer representation for the weights or activations. \cite{bit_efficient,jain2019trained}.
In this work, we apply pruning and quantization techniques for visual recognition tasks. We apply these  compression techniques on the convolutional neural networks which commonly comprise the backbone architectures of state-of-the-art recognition models such as in Faster R-CNN~\cite{fasterrcnn}. Model compression on CNNs is well suited as recent studies \cite{pas2020modeling} have shown that CNN structures have redundant parameters which can be removed without loss in performance. In addition to pruning, we further improve the efficiency of object recognition models by quanitizing their parameters from 32-bit float to 8-bit integer precision. Finally, we show that our compression techniques also support pre-traiend model weight initialization thus, making it possible to take advantage of published model zoos. 

\section{Methods and Experiments}
 Our approach compresses the model parameters by introducing sparsity into the weights and using few bits to represent each parameter with marginal performance sacrifice. 

\textbf{Pruning.}
We remove the parts of the model that contribute less or nothing to the final performance. \emph{Weight pruning} ranks the individual parameters in the parameter matrix $w$ according to their magnitude (absolute value), and then set to zero the smallest k\% of the weights. \emph{Unit/Neuron pruning} sets entire columns in the parameter matrix to zero. This corresponds to completely removing a neuron. We use $L_2$-norm to rank the columns of the parameters matrix.\\
We cast pruning as the following optimization problem.
%  so far everything is perfect

\begin{equation}
\min _{w} L(w ; D)=\min _{w} \frac{1}{n} \sum_{i=1}^{n} l(w;x_i y_i) \hspace{0.5cm} \textrm{s.t.} \quad\|w\|_{0} \leq k
\end{equation}
where $w$ are our model parameters, $D = \{x_i,y_i\}_{i=1}^n$ is our dataset, $l(w; x, y)$ is our loss function and $k$ represents the desired sparsity level in the parameters.

\textbf{Experimental Setup.}
Before applying any compression techniques, using our custom dataset of 1309 instances that we collected from East African parks, we consider the training of a faster RCNN with a Resnet50 backbone and with FPN \cite{fpn}.
Following the common experimental setting in related work on network pruning in \cite{pas2020modeling}, we extended their approach for image classification to object detection with a Faster RCNN architecture. Our approach can be easily extended to other object detection models such as YOLO \cite{redmon2018yolov3,redmon2015look} among others. 
Our model has three main blocks; backbone, a proposal generator and ROI heads. The whole model has a total of 41.4 Millions of trainable parameters.
The backbone has more than 60\% of the total parameters, which makes it the most targeted block in our compression experiments. We take advantage of the publicly available state-of-the-art object detection models in Detectron2 and its model zoos \cite{wu2019detectron2,Detectron2018}. We use the Pytorch pruning \cite{NEURIPS2019_9015} and quantization libraries implemented in Pytorch.
We implement global pruning of k\% parameters and neurons and then we quantize our parameters to 8 bits.

\textbf{Experimental Results.} Regardless of the task, pruning imposes a trade-off between model efficiency and quality, with pruning increasing the former while (typically) decreasing the latter.

Table 1 shows a comparison of different methods of pruning.  For low sparsity our approaches outperforms even the dense baseline, which is in line with regularization properties of network pruning. On large models, pruning shows reasonable performance even with extremely high sparsity level. 

%  so for everything is perfect

\begin{center}
   \begin{tabular}{lccccccccc}
   \hline\\
\multicolumn{7}{c}{Pruned percentage in the
   backbone}\\ \\
    \cline { 2 - 10 } & 0\% & 10\% & 20\% & 30\% & 40\% & 50\%  & 70\% & 80\% & 90\% \\
    \hline 
    AP50 & 87.92  & 87.97 & 88.40 & 88.15 & 86.95  & 83.48  & 77.42 & 60.24 & 0.17\\
     \hline
    Memory(MBs) & 165.6  & 154.88  & 144.16  & 133.44  &  122.72  & 112.0 & 90.56   & 79.84  &  69.12\\
    \hline
    
    \hline\\ 
    \multicolumn{7}{c}{Pruned percentage in the ROI head}\\ \\
    \cline {2 -10} & 0\% & 10\% & 20\% & 30\% & 40\% & 50\%  & 70\% & 80\% & 90\% \\
    \hline
         AP50 & 87.92  & 87.97 & 88.40 & 88.15 & 86.95  & 83.48  & 77.42 & 60.24 & 0.17\\
    \hline
        Memory(MBs) & 165.6  & 160.0  & 154.4  & 148.8  &  143.2  & 137.6 & 126.4   & 120.8  &  115.2\\
    \hline
    
    \hline\\
   \multicolumn{10}{c}{Pruned percentage in the both backbone and the ROI head}\\ \\
    \cline {2 -10} & 0\% & 10\% & 20\% & 30\% & 40\% & 50\%  & 70\% & 80\% & 90\% \\
    \hline
         AP50 & 87.92  & 87.72 & 87.04 & 83.60 & 73.64  & 61.18  & 54.69 & 33.12 & 0.14\\
    \hline
        Memory(MBs) & 165.6  & 149.2  & 132.9  & 116.6  &  100.3  & 84.0 & 51.3   & 35.0  &  18.7\\
    \hline
    \end{tabular} 
\end{center}

\section{Conclusion}
We show that pruning and quantization techniques can efficiently compress object recognition models with little loss in performance. We can prune 40\% of the model with loss of a few points in average precision. The reduction in memory allows for efficient storage and enables deployment of object detectors on devices of lower computational capacity. 

\bibliographystyle{plain}
\bibliography{bibliography.bib}

\begin{thebibliography}{10}

\bibitem{Detectron2018}
Ross Girshick, Ilija Radosavovic, Georgia Gkioxari, Piotr Doll\'{a}r, and
  Kaiming He.
\newblock Detectron.
\newblock \url{https://github.com/facebookresearch/detectron}, 2018.

\bibitem{jain2019trained}
Sambhav~R. Jain, Albert Gural, Michael Wu, and Chris~H. Dick.
\newblock Trained quantization thresholds for accurate and efficient
  fixed-point inference of deep neural networks, 2019.

\bibitem{objects_detection_survey}
Licheng Jiao, Fan Zhang, Fang Liu, Shuyuan Yang, Lingling Li, Zhixi Feng, and
  Rong Qu.
\newblock A survey of deep learning-based object detection.
\newblock {\em IEEE Access}, 7:128837–128868, 2019.

\bibitem{fpn}
Tsung-Yi Lin, Piotr Dollár, Ross Girshick, Kaiming He, Bharath Hariharan, and
  Serge Belongie.
\newblock Feature pyramid networks for object detection, 2016.

\bibitem{liu2018rethinking}
Zhuang Liu, Mingjie Sun, Tinghui Zhou, Gao Huang, and Trevor Darrell.
\newblock Rethinking the value of network pruning, 2018.

\bibitem{bit_efficient}
Prateeth Nayak, David Zhang, and Sek Chai.
\newblock Bit efficient quantization for deep neural networks, 2019.

\bibitem{pas2020modeling}
Morteza~Mousa Pasandi, Mohsen Hajabdollahi, Nader Karimi, and Shadrokh Samavi.
\newblock Modeling of pruning techniques for deep neural networks
  simplification, 2020.

\bibitem{NEURIPS2019_9015}
Adam Paszke, Sam Gross, Francisco Massa, Adam Lerer, James Bradbury, Gregory
  Chanan, Trevor Killeen, Zeming Lin, Natalia Gimelshein, Luca Antiga, Alban
  Desmaison, Andreas Kopf, Edward Yang, Zachary DeVito, Martin Raison, Alykhan
  Tejani, Sasank Chilamkurthy, Benoit Steiner, Lu~Fang, Junjie Bai, and Soumith
  Chintala.
\newblock Pytorch: An imperative style, high-performance deep learning library.
\newblock In H.~Wallach, H.~Larochelle, A.~Beygelzimer, F.~d\textquotesingle
  Alch\'{e}-Buc, E.~Fox, and R.~Garnett, editors, {\em Advances in Neural
  Information Processing Systems 32}, pages 8024--8035. Curran Associates,
  Inc., 2019.

\bibitem{BNN_survey}
Haotong Qin, Ruihao Gong, Xianglong Liu, Xiao Bai, Jingkuan Song, and Nicu
  Sebe.
\newblock Binary neural networks: A survey.
\newblock {\em Pattern Recognition}, 105:107281, Sep 2020.

\bibitem{redmon2015look}
Joseph Redmon, Santosh Divvala, Ross Girshick, and Ali Farhadi.
\newblock You only look once: Unified, real-time object detection, 2015.

\bibitem{redmon2018yolov3}
Joseph Redmon and Ali Farhadi.
\newblock Yolov3: An incremental improvement, 2018.

\bibitem{fasterrcnn}
Shaoqing Ren, Kaiming He, Ross Girshick, and Jian Sun.
\newblock Faster r-cnn: Towards real-time object detection with region proposal
  networks, 2015.

\bibitem{wu2019detectron2}
Yuxin Wu, Alexander Kirillov, Francisco Massa, Wan-Yen Lo, and Ross Girshick.
\newblock Detectron2.
\newblock \url{https://github.com/facebookresearch/detectron2}, 2019.

\end{thebibliography}

\end{document}